%% file: main.tex
\documentclass[10pt,conference]{IEEEtran}
\IEEEoverridecommandlockouts
\usepackage{cite}
\usepackage{amsmath,amssymb,amsfonts}
\usepackage{algorithmic}
\usepackage{graphicx}
\usepackage{textcomp}
\usepackage{xcolor}
\usepackage[table]{xcolor}
\usepackage{booktabs}
\usepackage{multirow, multicol}
\usepackage{booktabs}
\usepackage{siunitx}
\usepackage{hyperref}
\usepackage{cleveref}
\usepackage{xurl}

\definecolor{noloraBG}{HTML}{EAF7EF} % very light mint
\definecolor{loraBG}{HTML}{F3EEFF}   % very light lavender

\def\BibTeX{{\rm B\kern-.05em{\sc i\kern-.025em b}\kern-.08em
    T\kern-.1667em\lower.7ex\hbox{E}\kern-.125emX}}
\begin{document}
\bstctlcite{IEEEexample:BSTcontrol}
\title{Quantization-Robust LLM Unlearning via Low-Rank Adaptation}

\author{
\IEEEauthorblockN{
João Vitor Boer Abitante$^{1,2}$,
Joana Meneguzzo Pasquali$^1$,
Luan Fonseca Garcia$^2$,
Ewerton de Oliveira$^{3}$,\\
Thomas da Silva Paula$^{3}$,
Rodrigo C. Barros$^{1,4}$,
Lucas S. Kupssinskü$^1$
}
\IEEEauthorblockA{
$^1$MALTA -- Machine Learning Theory and Applications Lab\\
School of Technology, Pontifícia Universidade Católica do Rio Grande do Sul\\
}
\IEEEauthorblockA{
$^2$Núcleo Avançado de Inteligência Artificial (NAIA)\\
School of Technology, Pontifícia Universidade Católica do Rio Grande do Sul\\}
\IEEEauthorblockA{
$^{3}$Brazil R\&D, HP Inc\\
}
\IEEEauthorblockA{
$^{4}$Kunumi Institute, Brazil\\
}
% \author{\IEEEauthorblockN{Anonymous Authors}
% \IEEEauthorblockA{Paper under double-blind review}

% \and
% \and
% \and
% \and
% \and
}

\maketitle

\begin{abstract}
Large Language Model (LLM) unlearning aims to remove targeted knowledge from a trained model, but practical deployments often require post-training quantization (PTQ) for efficient inference. However, aggressive low-bit PTQ can mask unlearning updates, causing quantized models to revert to pre-unlearning behavior. We show that standard full-parameter fine-tuning often induces parameter changes that are too small to survive 4-bit quantization. We propose quantization-robust unlearning via low-rank adaptation (LoRA): we freeze the base model and concentrate unlearning into trainable adapters so that the effective update is preserved after quantization. On Llama-2-7B evaluated with MUSE dataset (BOOKS and NEWS), LoRA improves 4-bit utility by up to 7.93 points (NPO+GDR on BOOKS: 50.17 to 58.10) and yields higher 4-bit utility on NEWS for GA+GDR (40.06 to 44.82, increase of 4.76). LoRA also substantially reduces privacy leakage under 4-bit PTQ, e.g., for GA+KLR on BOOKS, PrivLeak moves from -25.68 to -5.86 (closer to ideal 0), while maintaining strong forgetting (VerMem and KnowMem near 0). Thus, using LoRA for Machine Unlearning is beneficial for scenarios where quantization is necessary for model deployment.

% Large Language Model (LLM) unlearning aims to remove targeted knowledge from a trained model, but practical deployments often require post-training quantization (PTQ) for efficient inference. However, aggressive low-bit PTQ can mask or erase unlearning updates, causing quantized models to revert to pre-unlearning behavior. We show that standard full-parameter fine-tuning often induce parameter changes that are too small to survive 4-bit quantization. We propose quantization-robust unlearning via low-rank adaptation (LoRA): we freeze the base model and concentrate unlearning into trainable adapters so that the effective update is preserved after quantization. On Llama-2-7B evaluated with MUSE dataset (BOOKS and NEWS), LoRA improves 4-bit utility by up to 7.93 points (NPO+GDR on BOOKS: 50.17 to 58.10) and yields higher 4-bit utility on NEWS for GA+GDR (40.06 to 44.82, an increase of 4.76). LoRA also substantially reduces privacy leakage under 4-bit PTQ, e.g., for GA+KLR on BOOKS, PrivLeak moves from -25.68 to -5.86 (closer to the ideal 0), while maintaining strong forgetting (VerMem and KnowMem near 0). Thus, we claim that using LoRA for Machine Unlearning is beneficial for scenarios where quantization is necessary for model deployment.
\end{abstract}


\input{01-INTRO}
\input{02-BACKGROUND}
\input{03-FAILURE_EXPLANATION}
\input{04-ROBUST_UNELARNING_VIA_LORA}

\input{05-EXPERIMENTAL_SETUP}
\input{06-RESULTS}
\input{07-CONCLUSION}
\input{08-ACKNOWLEDGMENTS}

\bibliographystyle{IEEEtran}
\bibliography{refs}

\end{document}

%% file: 01-INTRO.tex
\section{Introduction}
\label{sec:introduction}

% Large Language Models (LLMs) have shown unprecedented capabilities in natural language understanding and generation, yet they rely on training datasets that often contain sensitive, private, or copyrighted information. 
Large Language Models (LLMs) show strong natural language capabilities, but their training data often includes sensitive, private, or copyrighted content.
As a result, Machine Unlearning has emerged as a critical requirement to address data privacy regulations and to mitigate the retention of hazardous knowledge \cite{liu2024rethinkingmachineunlearninglarge}.
%The objective of unlearning is to erase specific subsets of data, \textit{forget set}, while preserving the model's utility on the remaining \textit{retain set}.

Current unlearning methods, such as Gradient Ascent (GA) and Negative Preference Optimization (NPO), typically operate by directly optimizing a Loss Function on the forget set while regularizing to maintain general capabilities ~\cite{zhang2024negativepreferenceoptimizationcatastrophic}.
These methods are effective in high-precision settings such as FP16 or BF16. However, LLM deployment in resource-constrained environments increasingly relies on quantization, which reduces numerical precision to lower memory use and improve throughput \cite{Xu2024ASO}.

%Recent research has uncovered a critical vulnerability at the intersection of these two fields: \textit{catastrophic failure of unlearning via quantization} \cite{zhang2025catastrophicfailurellmunlearning}. 
Recent research shows that post-training quantization (PTQ) can revert models to pre-unlearning state ~\cite{zhang2025catastrophicfailurellmunlearning}.
This phenomenon occurs because standard unlearning algorithms produce small weight updates that fail to cross the decision boundaries of coarse quantization grids. 
Specifically, in 4-bit quantization regimes, the discretization step size often exceeds the magnitude of the unlearning update, \textit{masking} the changes and recovering the forgotten knowledge \cite{zhang2025catastrophicfailurellmunlearning}.

To address this limitation, we propose a new approach: \textit{Quantization-Robust Unlearning via Low-Rank Adaptation (\textsc{LoRA})}. 
Unlike full-parameter unlearning, which distributes small, diffuse updates across the entire network, we hypothesize that restricting optimization to a low-rank subspace concentrates the unlearning signal, making the weight updates sufficiently large to be robust to quantization.
By freezing the pre-trained weights and training low-rank adapters, our work shows two key mechanisms to maintain unlearning after PTQ: 
(1) \textit{Optimization Dynamics}, enabling significantly higher learning rates without destroying general utility \cite{hu2021loralowrankadaptationlarge}.
(2) \textit{Magnitude Control via Architecture}: while higher learning rates in full-parameter fine-tuning (Full-FT) can bias the model towards the retain set \cite{zhang2025catastrophicfailurellmunlearning}, LoRA's explicit layer selection helps preserve utility \cite{biderman2024loralearnsforgets}.

In this work, we evaluate our approach in the MUSE benchmark~\cite{shi2024muse} with the Llama-2-7B model \cite{touvron2023llama}. Addressing the failure modes of standard unlearning algorithms highlighted by \cite{zhang2025catastrophicfailurellmunlearning}, we demonstrate that explicitly merging trained LoRA adapters \cite{hu2021loralowrankadaptationlarge} prior to quantization ensures that unlearning effects persist even in aggressive 4-bit formats.

% Our main contributions are summarized as follows: 
% \begin{itemize} 
%     \item We analyze the conflict between minimal weight updates and PTQ that leads to unlearning failures. 
%     \item We propose an unlearning framework that utilizes rank constraints and scaling factors to generate structural updates resilient to quantization noise. 
%     \item We provide empirical evidence that our work outperforms Full-FT in preserving unlearning after PTQ.
% \end{itemize} 

Our main contributions are: (i) analyzing the conflict between minimal weight updates and PTQ that causes unlearning failures; (ii) proposing an unlearning framework that uses rank constraints and scaling factors to generate structural updates robust to quantization noise; and (iii) showing empirically that our method outperforms Full-FT in preserving unlearning after PTQ.

% \begingroup
% \renewcommand\thefootnote{}%
% \footnotetext{\footnotesize Code available at: \url{https://anonymous.4open.science/r/Quantization-Robust-LoRA-Unlearning-B206/}}%
% \addtocounter{footnote}{-1}%
% \endgroup

\begingroup
\renewcommand\thefootnote{}% 
\footnotetext{\footnotesize Code available at: \\\url{https://github.com/JoaoVitorBoer/Quantization-Robust-LoRA-Unlearning}}%
\addtocounter{footnote}{-1}%
\endgroup

%% file: 02-BACKGROUND.tex
\section{Background}
\label{sec:background}

\subsection{Machine Unlearning in LLMs}
\label{subsec:mu_in_llms}
Machine unlearning is an option for addressing data privacy regulations, copyright concerns, and the removal of hazardous knowledge in LLMs \cite{liu2024rethinkingmachineunlearninglarge}. Formally, let $f_{\text{target}}$ denote a pre-trained model parameterized by $\theta$, initially trained on a dataset $\mathcal{D}_{\text{train}}$. We define the \textbf{forget set} $\mathcal{D}_{\text{forget}} \subset \mathcal{D}_{\text{train}}$ as the specific subset of data to be removed, and the \textbf{retain set} $\mathcal{D}_{\text{retain}} = \mathcal{D}_{\text{train}} \setminus \mathcal{D}_{\text{forget}}$ as the data whose knowledge must be preserved.

% The goal of an unlearning algorithm $U$ is to produce an unlearned model $f_{\text{unlearn}} = U(f_{\text{target}}, \mathcal{D}_{\text{forget}}, \mathcal{D}_{\text{retain}})$ that approximates the behavior of a model trained from scratch solely on $\mathcal{D}_{\text{retain}}$. While exact unlearning (retraining from scratch) provides the theoretical gold standard, it is often computationally prohibitive for LLMs. Consequently, current research focuses on approximate unlearning, which operates under a dual objective:

% \begin{enumerate}
%     \item \textbf{Forgetting:} Effectively eliminating the influence of $\mathcal{D}_{\text{forget}}$ such that the model cannot recall or reproduce the targeted information.
%     \item \textbf{Utility Preservation:} Maintaining performance on $\mathcal{D}_{\text{retain}}$ and generalizing to disjoint hold-out sets.
% \end{enumerate}

%The goal of an unlearning algorithm $U$ is to produce $f_{\text{unlearn}} = U(f_{\text{target}}, \mathcal{D}_{\text{forget}}, \mathcal{D}_{\text{retain}})$ that approximates a model trained solely on $\mathcal{D}_{\text{retain}}$. Since exact retraining is computationally prohibitive, approximate unlearning targets a dual objective: (1) \textbf{Forgetting}, which eliminates the influence of $\mathcal{D}_{\text{forget}}$; and (2) \textbf{Utility Preservation}, which maintains performance on $\mathcal{D}_{\text{retain}}$ and generalizes to hold-out sets.

The goal of an unlearning algorithm $U$ is to produce $f_{\text{unlearn}} = U(f_{\text{target}}, \mathcal{D}_{\text{forget}}, \mathcal{D}_{\text{retain}})$ that approximates a model trained solely on $\mathcal{D}_{\text{retain}}$. 
Since full retraining is prohibitive for LLMs, approximate unlearning methods target two competing objectives: (1) forgetting the influence of $D_{forget}$, and (2) preserving utility on $D_{retain}$ and unseen data.
%Since full retraining is computationally prohibitive for LLMs, approximate unlearning methods target two competing objectives: (1) \textbf{Forgetting} to eliminate the influence of $\mathcal{D}_{\text{forget}}$, and (2) \textbf{Utility Preservation} to maintain performance on $\mathcal{D}_{\text{retain}}$ and generalize to unseen data.

These competing objectives are typically balanced through the following optimization formulation:
$\mathbb{E}_{(x,y)\sim \mathcal{D}_f}\!\left[\mathcal{L}_{\text{forget}}(y\mid x;\theta)\right]
+\lambda\,\mathbb{E}_{(x,y)\sim \mathcal{D}_r}\!\left[\mathcal{L}_{\text{retain}}(y\mid x;\theta)\right]$, where $\mathcal{D}_f$ and $\mathcal{D}_r$ denote the forget and retain sets respectively, $\mathcal{L}_{\text{forget}}$ is a loss function that penalizes the retention of information from $\mathcal{D}_f$, $\mathcal{L}_{\text{retain}}$ is a loss function that ensures utility is preserved on $\mathcal{D}_r$, and $\lambda > 0$ is a regularization hyperparameter that balances these competing objectives.
    
% These competing objectives are typically balanced through the following optimization formulation:

% \begin{equation}
% \label{eq:unlearning_obj}
% \min_{\theta} \mathbb{E}_{(x,y)\sim \mathcal{D}_f} [\mathcal{L}_{\text{forget}}(y|x; \theta)] + \lambda \cdot \mathbb{E}_{(x,y)\sim \mathcal{D}_r} [\mathcal{L}_{\text{retain}}(y|x; \theta)]
% \end{equation}

%where $\mathcal{D}_{f}$ and $\mathcal{D}_{r}$ denote %samples drawn from 
%the forget and retain sets respectively, $\mathcal{L}_{\text{forget}}$ penalizes the retention of information from the forget set, $\mathcal{L}_{\text{retain}}$ ensures utility is preserved on the retain set, and $\alpha$ is a regularization hyperparameter balancing these competing objectives.

We study two forgetting objectives, GA and NPO, each combined with retain-set regularization.

% In this work, %we assess effective unlearning methods for LLMs by examiningtwo primary algorithmic families: \textbf{Gradient Ascent (GA)} and \textbf{Negative Preference Optimization (NPO)} and their integration with utility preservation strategies.
% we focus on two primary families of unlearning algorithms and study their integration with utility preservation.

% \textbf{Gradient Ascent (GA):} is an %fundamental
% unlearning strategy that effectively inverts the standard training objective, i.e., it explicitly minimizes the likelihood of the data within the forget set by ascending the gradient of the loss function and pushing the model away from patterns learned in the set~\cite{jang2023}. Because this imposed divergence is often unbounded, GA frequently results in catastrophic collapse where the model's general capabilities are severely degraded~\cite{liu2024rethinkingmachineunlearninglarge}.

\textbf{Gradient Ascent (GA):} is an unlearning strategy that inverts the standard training objective by minimizing the likelihood of data in the forget set, pushing the model away from patterns learned from that set~\cite{jang2023}. Because this divergence is often unbounded, GA can lead to catastrophic collapse, severely degrading the model's general capabilities~\cite{liu2024rethinkingmachineunlearninglarge}.

\textbf{Negative Preference Optimization (NPO):} To mitigate the instability of GA, NPO adapts the Direct Preference Optimization (DPO) framework by treating the forget set as negative preference data \cite{zhang2024negativepreferenceoptimizationcatastrophic}. Unlike GA, NPO incorporates the original pre-trained model $\theta_{\text{ref}}$ as a reference to bound the unlearning process. The loss function is derived as:
\begin{equation}\label{eq:loss_npo}
\mathcal{L}_{\text{NPO}}(\theta)
= -\frac{2}{\beta}\,\mathbb{E}_{(x,y)\sim \mathcal{D}_f}
\left[
\log \sigma\!\left(
-\beta \log \frac{P_\theta(y\mid x)}{P_{\theta_{\text{ref}}}(y\mid x)}
\right)
\right],
\end{equation}

where $\beta$ is a scaling factor (inverse temperature). This formulation effectively reweights the gradient updates: it applies stronger penalties to samples where the current model still retains high probability relative to the reference, while vanishing for samples effectively unlearned \cite{zhang2024negativepreferenceoptimizationcatastrophic}. This mechanism helps prevent the model from diverging too far from the reference distribution, thereby offering better stability than GA.

\subsubsection{Utility Preservation Strategies}
Since $\mathcal{L}_{\text{GA}}$ and $\mathcal{L}_{\text{NPO}}$ focus solely on the forget set, they do not guarantee the preservation of general knowledge. To address this, we use two regularization strategies on the retain set $\mathcal{D}_r$~\cite{Maini2024TOFUAT}:

% \textbf{Gradient Descent on Retain Set (GDR):} This strategy explicitly maintains performance on relevant knowledge by adding a standard cross-entropy objective on the retain set. The loss acts as a counter-balance to the unlearning update:
% \begin{equation}\label{eq:loss_gdr}
%     \mathcal{L}_{\text{GDR}}(\theta) = -\mathbb{E}_{(x, y) \sim \mathcal{D}_r} [\log P_\theta(y | x)].
% \end{equation}
% Combining this with a forgetting objective (e.g., GA+GDR) ensures the model continues to optimize for correct predictions on the retained data.

\textbf{Gradient Descent on Retain Set (GDR):} This strategy explicitly maintains utility by adding a cross-entropy objective on the retain set, defined as $\mathcal{L}_{\text{GDR}}(\theta) = -\mathbb{E}_{(x, y) \sim \mathcal{D}_r} [\log P_\theta(y | x)]$, which acts as a counter-balance to the unlearning update. Combining this with a forgetting objective (e.g., GA+GDR) ensures the model continues to optimize for correct predictions on the retained data.

\textbf{KL Minimization on Retain Set (KLR):} Alternatively, KLR preserves utility by minimizing the Kullback-Leibler divergence $\mathcal{L}_{\text{KLR}}(\theta) = \mathbb{E}_{x \sim \mathcal{D}_r} [ D_{\text{KL}}(P_{\theta_{\text{ref}}}(\cdot | x) \,||\, P_\theta(\cdot | x)) ]$, enforcing the unlearned model's output distribution to remain close to the original. This soft constraint prevents behavioral drift on $\mathcal{D}_r$ during updates~\cite{Maini2024TOFUAT, zhang2024negativepreferenceoptimizationcatastrophic}.

% \textbf{KL Minimization on Retain Set (KLR):} Alternatively, KLR preserves utility by enforcing that the output distribution of the unlearned model remains close to that of the original model on the retain set. It minimizes the Kullback-Leibler (KL) divergence:
% \begin{equation}\label{eq:loss_klr}
%     \mathcal{L}_{\text{KLR}}(\theta) = \mathbb{E}_{x \sim \mathcal{D}_r} \left[ D_{\text{KL}}(P_{\theta_{\text{ref}}}(\cdot | x) \,||\, P_\theta(\cdot | x)) \right].
% \end{equation}
% This soft constraint ensures that the model's behavior on $\mathcal{D}_r$ does not drift significantly during the unlearning updates \cite{Maini2024TOFUAT, zhang2024negativepreferenceoptimizationcatastrophic}.

In our experiments, we evaluate the performance of GA and NPO, as well as their regularized variants (GA+GDR, GA+KLR, NPO+GDR, and NPO+KLR), to analyze the trade-off between unlearning and utility preservation.

\subsection{LLM Quantization}
Quantization is a model compression technique that reduces the numerical precision of an LLM's parameters and activations, typically from high-precision floating-point formats (e.g., 32-bit) to lower-precision integer representations (e.g., 8-bit, 4-bit, or lower). The core trade-off is efficiency versus accuracy: fewer bits reduce storage and bandwidth use, but increase approximation error and can hurt perplexity or task performance \cite{Xu2024ASO}. There are two primary paradigms for quantization: Quantization-Aware Training (QAT), which simulates low-precision effects during training to allow the model to adapt, and Post-Training Quantization (PTQ), which converts a pre-trained model directly without extensive retraining.

\subsection{Low-Rank Adaptation (\textsc{LoRA})}

\textsc{LoRA} is a parameter-efficient fine-tuning method proposed to adapt LLMs to downstream tasks without the computational cost of Full-FT~\cite{hu2021loralowrankadaptationlarge}. Formally, for a pre-trained weight matrix $W_0 \in \mathbb{R}^{d \times k}$, \textsc{LoRA} freezes $W_0$ and constrains the weight update $\Delta W$ by representing it as a low-rank decomposition $W_0 + \Delta W = W_0 + BA$, where $B \in \mathbb{R}^{d \times r}$ and $A \in \mathbb{R}^{r \times k}$ are trainable matrices, and the rank $r \ll \min(d, k)$. 

In Machine Unlearning, \textsc{LoRA} can help reduce forgetting of the base model's capabilities compared to full fine tuning \cite{biderman2024loralearnsforgets}. This characteristic is particularly valuable in unlearning scenarios where the goal is to selectively forget specific knowledge while preserving the model’s general capabilities.

%% file: 03-FAILURE_EXPLANATION.tex
\section{Unlearning Failure via Quantization}
\label{sec:quant_fail}

Recent empirical observations indicate that while unlearning methods appear successful in full precision, unlearning effects are frequently erased upon quantization. This section provides a theoretical explanation of this phenomenon, adhering to the framework established by \cite{zhang2025catastrophicfailurellmunlearning}, identifying the conflict between the minimal weight updates characteristic of current unlearning algorithms and the resolution limits of low-precision quantization.

\textbf{Minimal Weight Change Constraint.} In Full-FT, the optimizer must balance the forgetting of specific samples against the preservation of the entire parameter distribution. To avoid catastrophic forgetting of the retain set $\mathcal{D}_{retain}$, unlearning benchmarks such as MUSE \cite{shi2024muse} and TOFU \cite{Maini2024TOFUAT} typically require small learning rates (e.g., $\eta \approx 10^{-5}$ to $10^{-7}$). This results in diffuse, low-magnitude updates spread across all parameters. Consequently, the unlearned weights $W_{u}$ remains proximate to the original weights $W_0$ and, therefore, the update $\Delta W = W_{u} - W_0$ is minute.

% Standard unlearning objectives (e.g., GA, NPO) are typically constrained by utility preservation terms (GDR, KLR) or small learning rates to prevent catastrophic forgetting of general knowledge. Benchmarks such as MUSE \cite{shi2024muse} and TOFU \cite{Maini2024TOFUAT} require learning rates ($10^{-5}$ to $10^{-7}$) that are orders of magnitude smaller than those used in standard fine-tuning~\cite{zhang2025catastrophicfailurellmunlearning}. Consequently, the unlearned weight $W_{u}$ remains numerically proximate to the original weight $W_0$. The unlearning update $\Delta W = W_{u} - W_0$ is, therefore, minute.

\textbf{Quantization Masking.}
This minimal deviation becomes critical during PTQ. Considering a group or block of weights, the quantization function $Q(\cdot)$ maps continuous weights into a discrete set of indices within the range $\left[-2^{N-1},\,2^{N-1}-1\right]$, using a step size $s$. A weight $W$ is mapped to a quantized value $q_i = i s$ if it falls within the interval:

\begin{equation}
\label{eq:quant_interval}
\mathcal{I}_i = \left[ \left(i - \frac{1}{2}\right)s, \; \left(i + \frac{1}{2}\right)s \right)
\end{equation}

For the unlearning effect to persist in the quantized model, the update $\Delta W$ must shift the weight from its original interval $\mathcal{I}_i$ to a different interval. However, if the weight update does not cross a quantization bin boundary, i.e., $W_0$ and $W_{u}=W_0+\Delta W$ lie in the same interval $\mathcal{I}_i$, then the quantized index remains unchanged, so $Q(W_{u}) = Q(W_0)$. When this equality holds for the majority of parameters, the quantized unlearned model becomes the same as the quantized original model, resulting in the recovery of the forgotten knowledge \cite{zhang2025catastrophicfailurellmunlearning}.

\textbf{Impact of Bit-Width.}
The likelihood of this failure is dictated by the bit-width $N$, which defines the step size $s = \frac{\max(\left\lvert W \right\rvert)}{2^{N-1}}$.
\begin{itemize}
    \item \textbf{8-bit Quantization:} With $2^{7}=128$ intervals, $s_{\text{int8}}$ is small, providing a resolution that can often capture the subtle shifts $\Delta W$ induced by unlearning. Thus, it maintains comparable performance to full-precision models \cite{zhang2025catastrophicfailurellmunlearning}.
    \item \textbf{4-bit Quantization:} With only $2^{3}=8$ intervals, the step size $s_{\text{int4}}$ increases (e.g., $\approx 16\times$ larger than $s_{\text{int8}}$).
\end{itemize}
Since the unlearning updates $\Delta W$ generated by regularized GA or NPO are typically smaller than the coarse $s_{\text{int4}}$, 4-bit quantization aggressively masks these changes. This theoretical threshold explains the empirical evidence in \cite{zhang2025catastrophicfailurellmunlearning}, where 4-bit quantization is observed to be catastrophic for unlearning, effectively reverting the model to its pre-unlearning state.

%% file: 04-ROBUST_UNELARNING_VIA_LORA.tex
\section{Robust Unlearning via LoRA}
\label{sec:robust_unlearning_lora}

To address the failure of unlearning under quantization described in \cref{sec:quant_fail}, we propose \textit{Quantization-Robust Unlearning via Low-Rank Adaptation (LoRA)}. While standard unlearning methods typically operate on the full parameter space, often resulting in minute weight updates that are erased by quantization, we hypothesize that restricting the unlearning optimization to a low-rank subspace concentrates the gradient signal, producing structural updates robust to the discretization noise of low-precision formats.

\textbf{Unlearning Formulation with LoRA.} Let $f_{\theta}$ be the target LLM with pre-trained weights $W_0 \in \mathbb{R}^{d \times k}$. In the standard unlearning setting described in \cref{subsec:mu_in_llms}, the optimization is performed over the full set of parameters $\theta = \{W_0\}$. In our proposed method, we freeze the pre-trained weights $W_0$ and introduce trainable low-rank matrices $B \in \mathbb{R}^{d \times r}$ and $A \in \mathbb{R}^{r \times k}$, where $r \ll \min(d, k)$ \cite{hu2021loralowrankadaptationlarge}. The forward pass for a layer becomes $h = W_0 x + \frac{\alpha}{r} BA x$, where $\alpha$ is a scaling hyperparameter constant in $r$. The unlearning objective function $\mathcal{L}_{total}$ is minimized solely with respect to the adapter parameters $\Phi = \{A, B\}$. By freezing $W_0$, we ensure that the base knowledge of the model is structurally preserved, shifting the unlearning burden entirely to the additive term $\Delta W = \frac{\alpha}{r}BA$ \cite{hu2021loralowrankadaptationlarge}.

As discussed in \cref{sec:quant_fail}, the primary cause of unlearning failure in quantized models is the ``Minimal Weight Change Constraint'' \cite{zhang2025catastrophicfailurellmunlearning}, where the unlearning update $\Delta W$ is smaller than the quantization step size $s$. We argue that \textsc{LoRA} overcomes this through two mechanisms: \textit{Optimization Dynamics and Step Size} and \textit{Magnitude Control via Scaling and Architecture}.

\textbf{Optimization Dynamics and Step Size.} As prior research has pointed out \cite{biderman2024loralearnsforgets}, LoRA imposes a low-rank constraint that serves as an implicit regularizer. Because the optimization is restricted to a subspace of rank $r$, the risk of distorting the model's general features is significantly reduced compared to Full-FT. This structural stability allows us to employ significantly larger learning rates (e.g., $\eta \approx 10^{-4}$) \cite{hu2021loralowrankadaptationlarge}, resulting in larger numerical updates within the targeted subspace.

% Full-FT unlearning is inherently bound by the minimal-update regime described in \cref{sec:quant_fail}. Due to the necessity of small learning rates, the resulting weight shifts $\|\Delta W\|_{\infty}$ frequently fail to surpass the quantization step size $\Delta$, causing the unlearning effect to be masked by PTQ.

% In full-parameter unlearning (Full-FT), the optimizer must balance the forgetting of specific samples against the preservation of the entire parameter distribution. To avoid catastrophic forgetting of the retain set $\mathcal{D}_{retain}$, Full-FT typically requires small learning rates (e.g., $\eta \approx 10^{-5}$ to $10^{-7}$). This results in diffuse, low-magnitude updates spread across all parameters. Because the maximum update magnitude $\|\Delta W\|_{\infty}$ is smaller than the quantization step size $\Delta$, these changes fail to cross decision boundaries and are effectively masked by quantization.

Crucially, this higher learning rate translates into a larger \textit{effective step size} for the weight updates. By taking larger optimization steps, the accumulated values in matrices $A$ and $B$ rapidly grow large enough to push the effective weight update $\Delta W$ across the quantization boundary. The higher learning rate ensures that the unlearning signal is not just a theoretical gradient direction, but a numerical displacement large enough to survive the quantization process.

\textbf{Magnitude Control via Scaling and Architecture.} Beyond the optimizer step size, LoRA makes unlearning robust to quantization with regard to the scaling factor $\alpha$ and layer selection. The scaling factor $\alpha$ acts as a direct amplifier of this signal. By tuning $\alpha$, we linearly scale the magnitude of the updates independent of the learning rate. This allows us to enforce the quantization threshold condition.

While increasing the learning rate in Full-FT might generate weight updates large enough to cross quantization boundaries, applying such large rates to the entire parameter set is risky. It can introduce a bias toward the retain data, skewing the model's behavior and degrading performance on disjoint tasks \cite{zhang2025catastrophicfailurellmunlearning}. To mitigate these side effects, we adopt a targeted strategy, akin to localized unlearning approaches \cite{lee2025doeslocalizationinformunlearning} by utilizing LoRA's capacity for explicit layer selection. Rather than distributing the unlearning budget across all layers, we target specific modules (e.g., MLP layers, attention projections or both) where knowledge is localized. This concentration of the unlearning objective not only preserves utility by limiting the scope of updates but also forces the update magnitude in those specific layers to be significantly higher to minimize the loss.

Consequently, the magnitude of the LoRA unlearning matrix updates is enough to persist after quantization, minimizing the masking effect common in full-FT methods.

% \textbf{Crossing the Quantization Threshold.} For the unlearning effect to persist after Post-Training Quantization (PTQ), the update must shift the weight index. Recall from (\ref{eq:quant_interval}) that a weight $w$ remains in the same quantization bin $\mathcal{I}_i$ if the update is insufficient to cross the boundary.

% In our method, the effective weight update is $\Delta W = \frac{\alpha}{r} B A$. Unlike the diffuse noise of Full-FT, the product $BA$ concentrates the unlearning signal into specific directions. Furthermore, the scaling factor $\alpha$ (typically $\alpha \ge r$) linearly amplifies the magnitude of these updates. By tuning $\alpha$, we can enforce the condition:

% \begin{equation}
% | \Delta W_{ij} | > \frac{\Delta}{2}
% \label{eq:quant_threshold}
% \end{equation}

% where $\Delta$ is the quantization step size defined in Section II. When this inequality holds for the salient weights involved in the ``forget'' knowledge, the quantized value shifts ($Q(W_{unlearn}) \neq Q(W_{ref})$), ensuring that the unlearning behavior is burned into the discrete representation of the model even at low bit-widths (e.g., 4-bit).

% Thus, LoRA transforms the unlearning problem from a \textit{diffuse, low-magnitude} perturbation (vulnerable to quantization) to a \textit{concentrated, high-magnitude} structural edit (robust to quantization).

%% file: 05-EXPERIMENTAL_SETUP.tex
\section{Experimental Setup}\label{sec:experimental_setup}

\input{Tables/01-BASELINES_FAILURE}

To evaluate the effectiveness and robustness of the proposed unlearning method, we utilize the Machine Unlearning Six-way Evaluation (MUSE) benchmark \cite{shi2024muse}. MUSE provides a framework for assessing unlearning across varying domains. We conduct experiments on two primary textual corpora provided by the benchmark:

\begin{itemize}
    \item \textbf{News:} This dataset comprises BBC news articles. It is partitioned into a \textit{forget set} (articles to be unlearned), a \textit{retain set} (articles to be preserved), and a \textit{holdout set} (for evaluating generalization) %\cite{shi2024muse}.
    \item \textbf{Books:} This dataset focuses on the Harry Potter series. The \textit{forget set} consists of the original novel texts, while the \textit{retain set} includes related content from the Harry Potter FanWiki. This split is designed to test the model's ability to unlearn specific verbatim content while retaining domain-related knowledge %\cite{shi2024muse}.
\end{itemize}

For both corpora, the benchmark provides two data formats: \textit{Verbatim} text (raw sequences for evaluating verbatim memorization) and \textit{Knowledge} sets (generated question-answer pairs) to assess the removal of semantic knowledge.

\subsection{Evaluation Metrics} Following the MUSE protocol \cite{shi2024muse}, we assess performance using four key metrics that balance the trade-off between forgetting, utility, and privacy:

\textbf{Verbatim Memorization (VerMem).} Measures the model's tendency to reproduce the forget set verbatim. The model is prompted with the first $l$ tokens from a sequence $x[:l]$ from the forget set $\mathcal{D}_f$, and the generated continuation is compared to the ground truth $x[l+1:]$ using the ROUGE-L F1 score, calculated as $\text{VerMem}(f, \mathcal{D}_f) = \frac{1}{|\mathcal{D}_f|} \sum_{x \in \mathcal{D}_f} \text{ROUGE}(f(x[:l]), x[l+1:])$. Lower scores indicate better unlearning.

\textbf{Knowledge Memorization (KnowMem).} Evaluates if the model retains semantic knowledge of the forgotten data. It computes the ROUGE-L score between the model's answer $f(q)$ and the ground truth answer $a$ for QA pairs in the forget set $\mathcal{D}_f$, defined as
$\text{KnowMem}(f,\mathcal{D}_f)=\frac{1}{|\mathcal{D}_f|}\sum_{(q,a)\in\mathcal{D}_f}\text{ROUGE}(f(q),a)$.
Lower scores indicate effective knowledge erasure.

\textbf{Privacy Leakage (PrivLeak).} Assesses the indistinguishability between the
unlearned model and a retrained model using Membership Inference Attacks (MIA). It
uses the Min-K \% Prob method to compute the AUC-ROC of discriminating between
$\mathcal{D}_f$ and $\mathcal{D}_r$. The metric is defined as the relative degradation
compared to a model retrained from scratch ($f_{\mathrm{retrain}}$):
$\text{PrivLeak}=
(\mathrm{AUC}(f_{\mathrm{unlearn}}\allowbreak;\mathcal{D}_{\mathrm{f}},\mathcal{D}_{\mathrm{r}})
-\mathrm{AUC}(f_{\mathrm{retrain}}\allowbreak;\mathcal{D}_{\mathrm{f}},\mathcal{D}_{\mathrm{r}}))
\allowbreak\;/\;\allowbreak
\mathrm{AUC}(f_{\mathrm{retrain}}\allowbreak;\mathcal{D}_{\mathrm{f}},\mathcal{D}_{\mathrm{r}})$.
Optimal scores are near zero, indicating the unlearned model leaks no more information than a model that never saw the data.

\textbf{Utility Preservation (UtilityPres).} Ensures general capabilities are maintained. We measure this by computing the \textit{Knowledge Memorization} score (ROUGE-L) on the retain set $\mathcal{D}_r$. Higher scores indicate better preservation of general knowledge.

\subsection{Implementation Details}

%\textbf{Model and Baselines.}
% We employ Llama-2-7B as the base model for all experiments. We evaluate two primary families of unlearning algorithms: GA and NPO. To ensure fair comparison and utility preservation, both methods are coupled with regularization strategies: GDR and KLR. This results in six baseline configurations: GA, NPO, GA+GDR, GA+KLR, NPO+GDR, and NPO+KLR. Additionally, we pair these baselines with LoRA and compare them against their full-parameter fine-tuned counterparts.

We use Llama-2-7B for all experiments and evaluate GA and NPO, with and without GDR or KLR regularization, yielding six baselines: GA, NPO, GA+GDR, GA+KLR, NPO+GDR, and NPO+KLR. 
%\textbf{Proposed LoRA Unlearning Configuration.}
% To evaluate unlearning with \textsc{LoRA}, updates are maintained after PTQ. We freeze the pre-trained weights $W_0$ and inject trainable \textsc{LoRA} adapters into all linear layers, including MLP modules and Attention projections. 

% We performed a grid search over quantization-robustness hyperparameters, sweeping LoRA ranks $r\in\{16,32,64,128\}$, coupling the scaling factor to rank with $\alpha\in\{0.5r,\,r,\,2r\}$, and tuning optimization settings via learning rates $\eta\in\{10^{-4}, 7\times10^{-4}\}$ and training durations of $\{5,10\}$ epochs. For unlearning methods with KLR and GDR, we searched for the optimal regularization weight $\lambda\in\{0.1, 1, 2, 10, 50, 100, 200, 300\}$, and these weights were fixed for LoRA experiments to ensure that performance improvements are attributable solely to LoRA. We set the NPO $\beta=0.1$~ as done in \cite{zhang2024negativepreferenceoptimizationcatastrophic}.
To evaluate unlearning with \textsc{LoRA}, updates are maintained after PTQ. We freeze the pre-trained weights $W_0$ and inject trainable \textsc{LoRA} adapters into all linear layers, which was selected via a grid search over all linear layers, MLP-only modules, and Attention-only projections.

% We performed a grid search over quantization-robustness hyperparameters, sweeping LoRA ranks $r\in\{16,32,64,128\}$, coupling the scaling factor to rank with $\alpha\in\{0.5r,\,r,\,2r\}$, and tuning optimization settings via learning rates $\eta\in\{10^{-4}, 7\times10^{-4}\}$ and training durations of $\{5,10\}$ epochs. 
We performed a grid search over quantization-robustness hyperparameters, sweeping LoRA ranks $r \in \{16, 32, 64, 128\}$, scaling factors $\alpha \in \{0.5r, r, 2r\}$, learning rates $\eta \in \{10^{-4}, 7 \times 10^{-4}\}$, and training durations of $\{5, 10\}$ epochs. For unlearning methods with KLR and GDR, we searched for the optimal regularization weight $\lambda\in\{0.1, 1, 2, 10, 50, 100, 200, 300\}$, and these weights were fixed for LoRA experiments to ensure that performance improvements are attributable solely to LoRA. We set the NPO $\beta=0.1$~ as done in \cite{zhang2024negativepreferenceoptimizationcatastrophic}.

Crucially, for all LoRA-based experiments, we explicitly merge the trained low-rank adapters into the base model parameters before quantization. This ensures that the quantization step is applied to the final unlearned weights ($W_{unlearn} = W_0 + \Delta W$), thereby subjecting the unlearning updates to the potential masking effects described in \cref{sec:quant_fail}.

We employ Round-to-Nearest (RTN) as our primary post-training quantization method. We note that recent studies have demonstrated that advanced calibration-based methods, such as GPTQ and AWQ, exhibit similar failure modes at 4-bit precision due to the resolution limits discussed in \cite{zhang2025catastrophicfailurellmunlearning}. We report the degradation in unlearning metrics as the bit-width decreases across three settings: \textbf{BF16} (original bfloat16 precision), \textbf{Int8} (8-bit post-training quantization), and \textbf{Int4} (4-bit post-training quantization).

%% file: Tables/01-BASELINES_FAILURE.tex
\begin{table*}[t]
\caption{Unlearning performance of full-precision vs. quantized models on BOOKS and NEWS corpora from MUSE~\cite{shi2024muse}.}
\label{tab:baselines_failure}
\centering
\footnotesize
\setlength{\tabcolsep}{3pt}
\renewcommand{\arraystretch}{1.15}
\sisetup{
  table-number-alignment = center,
  table-format = -2.2
}

\begin{tabular}{@{}ll *{4}{S} *{4}{S}@{}}
\toprule
\multirow{2}{*}{\textbf{Method}} & \multirow{2}{*}{\textbf{Prec.}} &
\multicolumn{4}{c}{\textbf{BOOKS}} &
\multicolumn{4}{c}{\textbf{NEWS}} \\
\cmidrule(lr){3-6}\cmidrule(lr){7-10}
& &
\multicolumn{1}{c}{VerMem ($\downarrow$)} &
\multicolumn{1}{c}{KnowMem ($\downarrow$)} &
\multicolumn{1}{c}{PrivLeak ($\rightarrow 0$)} &
\multicolumn{1}{c}{UtilityPres ($\uparrow$)} &
\multicolumn{1}{c}{VerMem ($\downarrow$)} &
\multicolumn{1}{c}{KnowMem ($\downarrow$)} &
\multicolumn{1}{c}{PrivLeak ($\rightarrow 0$)} &
\multicolumn{1}{c}{UtilityPres ($\uparrow$)} \\
\midrule

\textbf{$f_{\text{target}}$} & Full  & 99.70 & 47.12 & -57.34 & 69.56 & 56.85 & 63.72 & -99.81 & 55.21 \\
                            & 8-bit & 99.70 & 49.74 & -57.38 & 64.22 & 57.36 & 67.28 & -99.81 & 56.60 \\
                            & 4-bit & 94.78 & 36.35 & -60.41 & 51.05 & 46.35 & 55.10 & -99.79 & 49.70 \\
% \addlinespace[2pt]
\midrule

\textbf{GA} & Full  & 0.00 & 0.00 & -19.00 & 0.00 & 0.00 & 0.00 & 51.97 & 0.00 \\
            & 8-bit & 0.00 & 0.00 & -18.97 & 0.00 & 0.00 & 0.00 & 52.53 & 0.00 \\
            & 4-bit & 0.00 & 0.00 & -19.14 & 0.00 & 0.00 & 0.00 & 53.27 & 0.00 \\
\addlinespace[2pt]

\textbf{GA+GDR} & Full  & 0.00 & 36.30 & -24.01 & 68.74 & 52.15 & 56.98	& -99.79 & 49.57 \\
                & 8-bit & 0.00 & 30.86 & -23.95 & 69.26 & 50.22	& 58.32 & -99.79 & 46.19 \\
                & 4-bit & 0.00 & 28.77 & -23.65 & 53.79 & 42.07 & 48.19 & -99.79 & 40.06 \\
\addlinespace[2pt]

\textbf{GA+KLR} & Full  & 0.00 & 34.62 & -24.66 & 62.14 & 49.01 & 63.12 & -99.51 & 52.14 \\
                & 8-bit & 0.00 & 35.73 & -24.74 & 59.61 & 48.50 & 63.29 & -99.53 & 52.19 \\
                & 4-bit & 0.00 & 23.64 & -25.68 & 44.13 & 43.38 & 53.24 & -99.51 & 44.18 \\
\addlinespace[2pt]

\textbf{NPO} & Full  & 13.00 & 5.25	& -55.67 & 10.60 & 15.52 & 37.30 & -84.82 & 35.61 \\
             & 8-bit & 12.53 & 5.11 & -55.58 & 9.97 & 15.43 & 35.54 & -84.84 & 35.23 \\
             & 4-bit & 13.33 & 6.65 & -56.63 & 12.41 & 15.89 & 37.51 & -85.57 & 34.25 \\
\addlinespace[2pt]

\textbf{NPO+GDR} & Full  & 54.61 & 33.39 & -56.37 & 60.09 & 26.89 & 52.11 & -86.04 & 48.90 \\
                 & 8-bit & 55.83 & 31.64 & -56.69 & 61.25 & 26.09 & 53.44 & -86.33 & 48.89 \\
                 & 4-bit & 41.18 & 25.64 & -58.45 & 50.17 & 23.91 & 47.63 & -87.53 & 44.01 \\
\addlinespace[2pt]

\textbf{NPO+KLR} & Full  & 51.39 & 31.16 & -55.82 & 60.25 & 24.03 & 45.81 & -86.85 & 48.13 \\
                 & 8-bit & 50.11 & 28.71 & -55.88 & 59.62 & 22.69 & 46.53 & -86.85 & 44.80 \\
                 & 4-bit & 38.65 & 26.00 & -57.87 & 48.50 & 22.09 & 46.80 & -87.63 & 44.76 \\
\bottomrule
\end{tabular}

% \vspace{2pt}
\footnotesize\emph{Note: $\downarrow$ lower is better, $\uparrow$ higher is better, and $\rightarrow 0$ closer to zero is better.}
\end{table*}

%% file: 06-RESULTS.tex
\section{Results} \label{sec:results}

%In this section, we present an empirical evaluation of our LoRA-based unlearning framework using the MUSE benchmark \cite{shi2024muse}. We begin by examining how post-training quantization affects standard full fine-tuning (Full-FT) unlearning methods (Table \ref{tab:baselines_failure}), corroborating the quantization masking effects discussed in Section \ref{sec:quant_fail}. We then study whether Low-Rank Adaptation (LoRA) can mitigate this degradation under 4-bit quantization (Table \ref{tab:lora_results}).

% \input{Tables/01-BASELINES_FAILURE}
\subsection{Failure of Full Fine-tuning Unlearning}
We first evaluate standard Full-FT unlearning baselines on Llama-2-7b. \Cref{tab:baselines_failure} compares full-precision (BF16) results against post-training quantized variants (Int8 and Int4).

From Table \ref{tab:baselines_failure}, we observe that most quantized models exhibit reduced performance across all metrics, with the most severe degradation occurring under 4-bit quantization. 
This behavior is consistent with the theoretical analysis in Section \ref{sec:quant_fail}, because many unlearning algorithms operate under small, utility-preserving updates, the induced parameter changes are often too small to survive the coarse discretization of Int4.

An exception is \textbf{GA}, which appears to achieve near-complete forgetting even after 4-bit quantization. However, this result is misleading: GA lacks an explicit utility-preservation constraint, and its apparent ``success'' stems from a near-complete collapse of model utility (Utility $\approx 0$).

In contrast, 8-bit quantization yields performance that is generally closer to full precision across methods. This aligns with our earlier discussion (Section \ref{sec:quant_fail}), Int8 provides finer quantization resolution and is therefore more sensitive to (and more likely to retain) the relatively small weight changes induced by utility-regularized unlearning.

Finally, these results highlight a practical constraint: methods without utility regularization can achieve low memorization metrics by substantially degrading utility, and are therefore not strong candidates for quantization-robust unlearning. Accordingly, in this study we apply \textsc{LoRA} only to objectives paired with explicit utility regularization (GDR or KLR). This choice is motivated by the observation that unconstrained objectives such as GA or NPO can induce excessive unlearning accompanied by utility degradation.

\subsection{Quantization-Robust Unlearning with LoRA}
\input{Tables/02-LORA_RESULTS}
We next investigate whether applying \textsc{LoRA} GA+GDR, GA+KLR, NPO+GDR and NPO+KLR preserve the unlearning signal after 4-bit PTQ. Table \ref{tab:lora_results} summarizes these results.

Overall, \textsc{LoRA} improves quantization robustness across utility-regularized unlearning methods, but the resulting trade-offs depend on the underlying objective and dataset. 
On BOOKS, \textsc{LoRA} often yields stronger forgetting signals on at least one memorization axis (particularly VerMem) and can substantially reduce privacy leakage (PrivLeak) toward the ideal $0$ for GA+\{GDR, KLR\}. 
We also highlight GA+KLR, for which \textsc{LoRA} can drive both VerMem and KnowMem close to $0$, keeping it stable even after 4-bit quantization.

A key benefit is improved \emph{robustness of utility} under Int4. For instance, for GA+GDR on BOOKS, although \textsc{LoRA} reduces full-precision utility (Utility $68.74 \rightarrow 61.90$), it makes the model less sensitive to 4-bit quantization: the utility drop is considerably smaller with \textsc{LoRA} ($61.90 \rightarrow 53.16$) than with Full-FT ($68.74 \rightarrow 53.79$). 
Similar robustness trends are observed on NEWS, where \textsc{LoRA} yields higher Int4 utility for GA+GDR ($40.06 \rightarrow 44.82$) and reduces the quantization-induced utility drop for GA+KLR ($52.29 \rightarrow 47.77$ vs. $52.14 \rightarrow 44.18$ for Full-FT).

For NPO with regularization, \textsc{LoRA} strengthens forgetting on BOOKS while maintaining stable utility under quantization. 
In particular, for NPO+GDR, \textsc{LoRA} improves VerMem forgetting relative to Full-FT and remains essentially unchanged from full precision to Int4 in both forgetting and utility (Utility $59.65 \rightarrow 58.10$), demonstrating improved quantization robustness compared to Full-FT (Utility $60.09 \rightarrow 50.17$). 
On NEWS, we observe similar robustness utility trends.

Similarly, for NPO+KLR, \textsc{LoRA} provides a highly quantization-stable on BOOKS, with all metrics remaining nearly unchanged between full precision and Int4 (e.g., VerMem $16.76 \rightarrow 17.03$, Utility $41.82 \rightarrow 42.02$). 
On NEWS, \textsc{LoRA} exhibits similar PTQ stability, although it does not consistently outperform Full-FT in absolute forgetting or utility.

Across methods, \textsc{LoRA} reduces the sensitivity of unlearning to Int4 PTQ and, in many cases, improves the unlearning outcome itself (e.g., stronger forgetting signals and reduced privacy leakage on BOOKS). However, the best operating point still depends on the desired balance between forgetting, privacy, and utility. In the most stable settings (e.g., NPO+KLR on BOOKS), metrics remain nearly unchanged between full precision and Int4, indicating robustness to aggressive quantization. Among the evaluated approaches, GA+KLR and GA+GDR with \textsc{LoRA} provide the clearest improvements, combining stronger forgetting/privacy gains with improved robustness.

%% file: Tables/02-LORA_RESULTS.tex
% \definecolor{noloraBG}{cmyk}{0.05, 0, 0, 0} % Light Cyan/Blue
% \definecolor{loraBG}{cmyk}{0, 0.05, 0.1, 0} % Light Apricot/Orange
\definecolor{noloraBG}{cmyk}{0.05, 0, 0, 0} % light mint
\definecolor{loraBG}{cmyk}{0, 0.05, 0.1, 0}

\begin{table*}[t]
\caption{Baseline unlearning results on BOOKS and NEWS with/without LoRA under full precision and 4-bit quantization.}
\label{tab:lora_results}
\centering
\footnotesize
\renewcommand{\arraystretch}{1.15}
\setlength{\tabcolsep}{3pt}

\sisetup{
  table-number-alignment = center,
  table-format = -2.2
}

\begin{tabular}{@{}lll *{4}{S} *{4}{S}@{}}
\toprule
\multirow{2}{*}{\textbf{Method}} &
\multirow{2}{*}{\textbf{Prec.}} &
\multirow{2}{*}{\textbf{Adapter}} &
\multicolumn{4}{c}{\textbf{BOOKS}} &
\multicolumn{4}{c}{\textbf{NEWS}} \\
\cmidrule(lr){4-7}\cmidrule(lr){8-11}
& & &
\multicolumn{1}{c}{VerMem ($\downarrow$)} &
\multicolumn{1}{c}{KnowMem ($\downarrow$)} &
\multicolumn{1}{c}{PrivLeak ($\rightarrow 0$)} &
\multicolumn{1}{c}{UtilityPres ($\uparrow$)} &
\multicolumn{1}{c}{VerMem ($\downarrow$)} &
\multicolumn{1}{c}{KnowMem ($\downarrow$)} &
\multicolumn{1}{c}{PrivLeak ($\rightarrow 0$)} &
\multicolumn{1}{c}{UtilityPres ($\uparrow$)} \\
\midrule

% --- GA+GDR GROUP ---
% Row 1: No LoRA
% \rowcolor{noloraBG} 
\textbf{GA+GDR} & Full  & -- & 0.00 & 36.30 & -24.01 & 68.74 & 52.15 & 56.98 & -99.79 & 49.57 \\
% Row 2: LoRA
% \rowcolor{loraBG}   
                & Full  & LoRA & 0.00 & 37.68 & -3.79 & 61.90 & 46.49 & 52.13 & -99.79 & 47.78 \\
% Row 3: No LoRA
% \rowcolor{noloraBG} 
                & 4-bit & -- & 0.00 & 28.77 & -23.65 & 53.79 & 42.07 & 48.19 & -99.79 & 40.06 \\
% Row 4: LoRA
% \rowcolor{loraBG} 
                & 4-bit & LoRA & 0.00 & 26.43 & -3.77 & 53.16 & 40.22 & 48.15 & -99.79 & 44.82 \\
\midrule
\addlinespace[2pt]

% --- GA+KLR GROUP ---
% \rowcolor{noloraBG}
\textbf{GA+KLR} & Full  & --   & 0.00 & 34.62 & -24.66 & 62.14 & 49.01 & 63.12 & -99.51 & 52.14 \\
% \rowcolor{loraBG}
                & Full  & LoRA & 0.00 & 0.00 & -3.67 & 62.19 & 52.33 & 60.11 & -99.74 & 52.29 \\
% \rowcolor{noloraBG}
                & 4-bit & --   & 0.00 & 23.64 & -25.68 & 44.13 & 43.38 & 53.24 & -99.51 & 44.18 \\
% \rowcolor{loraBG}
                & 4-bit & LoRA & 0.14 & 0.00 & -5.86 & 50.30 & 41.72 & 53.68 & -99.74 & 47.77 \\
\midrule
\addlinespace[2pt]

% --- NPO+GDR GROUP ---
% \rowcolor{noloraBG}
\textbf{NPO+GDR} & Full  & --   & 54.61 & 33.39 & -56.37 & 60.09 & 26.89 & 52.11 & -86.04 & 48.90 \\
% \rowcolor{loraBG}
                 & Full  & LoRA & 22.67 & 36.63 & -60.07 & 59.65 & 46.39 & 59.51 & -99.74 & 48.61\\
% \rowcolor{noloraBG}
                 & 4-bit & --   & 41.18 & 25.64 & -58.45 & 50.17 & 23.91 & 47.63 & -87.53 & 44.01 \\
% \rowcolor{loraBG}
                 & 4-bit & LoRA & 20.30 & 36.64 & -58.91 & 58.10 & 37.78 & 49.09 & -99.72 & 46.40 \\
\midrule
\addlinespace[2pt]

% --- NPO+KLR GROUP ---
% \rowcolor{noloraBG}
\textbf{NPO+KLR} & Full  & --   & 51.39 & 31.16 & -55.82 & 60.25 & 24.03 & 45.81 & -86.85 & 48.13 \\
% \rowcolor{loraBG}
                 & Full  & LoRA & 16.76 & 26.48 & -61.32& 41.82 & 35.67 & 48.30 & -94.73 & 40.89 \\
% \rowcolor{noloraBG}
                 & 4-bit & --   & 38.65 & 26.00 & -57.87 & 48.50 & 22.09 & 46.80 & -87.63 & 44.76 \\
% \rowcolor{loraBG}
                 & 4-bit & LoRA & 17.03 & 24.33  & -56.88 & 42.02 & 28.24 & 48.40 & -95.42 & 39.96 \\
\bottomrule
\end{tabular}

\end{table*}

%% file: 07-CONCLUSION.tex
\section{Conclusion}
\label{sec:conclusion}

This paper studied the failure of LLM Unlearning with PTQ, especially under aggressive 4-bit quantization. %Building on the ``quantization masking'' explanation, 
%We showed that Full-FT unlearning objectives often induce updates that are too small to survive PTQ, yielding quantized models that behave similarly to the pre-unlearning baseline.
To mitigate this failure, we proposed \emph{quantization-robust unlearning via LoRA}, which freezes the base model and concentrates unlearning into trainable low-rank adapters.

We found that merging LoRA adapters before PTQ substantially improves 4-bit robustness. Compared with Full-FT, LoRA-based unlearning preserves the forgetting/privacy signal after quantization and often reduces the drop in utility.
% We found that merging LoRA adapters before PTQ substantially improves robustness at 4-bit. 
% When compared to Full-FT, LoRA-based unlearning preserves the forgetting/privacy signal after quantization and, in many settings, reduces the quantization-induced drop in utility.
Our findings suggest that parameter-efficient, structurally constrained updates offer a path toward deployable unlearning in resource-constrained, low-precision regimes.% Future research should explore unlearning knowledge localization and the development of quantization-aware objectives resilient across various quantizers (e.g., GPTQ and AWQ) and diverse model architectures.

%% file: 08-ACKNOWLEDGMENTS.tex
\section*{Acknowledgment}\label{sec:ack}

This paper was achieved in a project supported by the Brazilian Informatics Law (Law nº 8.248 of 1991) and was developed over Agreement 001/2015 between Pontifícia Universidade Católica do Rio Grande do Sul and HP Brasil Indústria e Comércio de Equipamentos Eletrônicos Ltda.
This study was financed in part by the Coordination for the Improvement of Higher Education Personnel – Brazil (CAPES) – Finance Code 001.
This study was financed in part by Conselho Nacional de Desenvolvimento Científico e Tecnológico - Brazil - (CNPq) - Grant Number: 443072/2024-8.
This study was financed in part by Fundação de Amparo à Pesquisa do Estado do Rio Grande do Sul (FAPERGS) - Grant Number: 25/2551-0000891-3.
This work was supported by Kunumi Institute. The authors thank the institution for its financial support and commitment to advancing scientific research.

%Portions of this manuscript were refined with the assistance of Google Gemini for language review and editing. All technical content, analyses, interpretations, and conclusions are the author's own.

During the preparation of this work, the authors used Google Gemini in order to proofread the manuscript. The authors take full responsibility for the content of the publication.